# Forecasting Smog Events Using ConvLSTM: A Spatio-Temporal Approach for Aerosol Index Prediction in South Asia

Taimur Khan[1,2]

[1] Community Ecology, Helmholtz Centre for Environmental Research - UFZ, Theodor-Lieser-Straße 4, 06120 Halle (Saale), Germany – taimur.khan@ufz.de

[2] Faculty of Physics and Earth System Sciences; Leipzig University, Talstraße 35, 04103 Leipzig, Germany – tq89iwow@studserv.uni-leipzig.de

**Keywords**: air quality, sentinel, remote sensing, deep learning, forecasting, smog, south asia, smog season

**Corresponding Author:** Taimur Khan, Mail @ taimur.khan@ufz.de, Telephone @ +49 341 235 5532

**High-Resolution Figures:**
https://drive.google.com/drive/folders/1RUZqiAqrd_eLpjytHpLelwyAVKiDCPHz?usp=share_link

**Code:** https://git.sc.uni-leipzig.de/ss2024-12-geo-m-ds02/smogseer

**Data:** https://zenodo.org/records/13118498

# Abstract

The South Asian Smog refers to the recurring annual air pollution events marked by high contaminant levels, reduced visibility, and significant socio-economic impacts, primarily affecting the Indo-Gangetic Plains (IGP) from November to February. Over the past decade, increased air pollution sources such as crop residue burning, motor vehicles, and changing weather patterns have intensified these smog events. However, real-time forecasting systems for increased particulate matter concentrations are still not established at regional scale. The Aerosol Index, closely tied to smog formation and a key component in calculating the Air Quality Index (AQI), reflects particulate matter concentrations. This study forecasts aerosol events using Sentinel-5P air constituent data (2019-2023) and a Convolutional Long-Short Term Memory (ConvLSTM) neural network, which captures spatial and temporal correlations more effectively than previous models. Using the Ultraviolet (UV) Aerosol Index at 340-380 nm as the predictor, results show the Aerosol Index can be forecasted at five-day intervals with a Mean Squared Error of ~0.0018, loss of ~0.3995, and Structural Similarity Index of ~0.74. While effective, the model can be improved by integrating additional data and refining its architecture.

# Table of Contents

# 1. Introduction

### a. Background of smog & extreme smog events

In South Asia, the term "fifth season" refers to a period of severe air pollution, particularly characterised by extreme smog events, which typically occurs during the late autumn and early winter months (Guttikunda & Gurjar, 2012). This phenomenon is especially prominent in the northwestern regions of the Indo-Gangetic Plains (IGP), where air quality deteriorates significantly during this time. The primary contributors to this so-called "fifth season" include the burning of crop residue by farmers, vehicular emissions, industrial pollution, and dust from construction activities (Lelieveld et al., 2015; Bharali et al., 2023). These pollutants accumulate and are exacerbated by specific meteorological conditions, such as temperature inversions and low wind speeds, which trap pollutants near the ground. This leads to the formation of a dense layer of smog, reducing visibility and posing severe health risks to the population (Cusworth, 2018). While the issue of extreme smog events is regionally significant, efforts to forecast and mitigate these events have been largely fragmented at national or provincial levels, with limited collaborative initiatives at the broader regional scale (Amann et al., 2013).

### b. The human & environmental health element

Extreme smog events are episodes of severe air pollution where concentrations of particulate matter (PM2.5 and PM10) and pollutants like sulphur dioxide, nitrogen oxides, and carbon monoxide reach hazardous levels, leading to health crises, particularly among vulnerable populations such as children, the elderly, and those with preexisting conditions (Guttikunda & Gurjar, 2012). These events obscure sunlight, causing visibility issues and disruptions to daily life, while also contributing to long-term economic and health costs (Lelieveld et al., 2015). The environmental impacts are profound, as smog reduces photosynthesis by blocking sunlight, damaging plant health, and affecting entire food chains (Mishra & Shibata, 2012). Pollutants from smog can degrade soil and contaminate water bodies, leading to biodiversity loss, while contributing to acid rain, which harms forests, crops, and aquatic ecosystems (Sarkar et al., 2018; Gautam et al., 2023). Additionally, ground-level ozone present in smog causes direct damage to vegetation, further exacerbating ecological disruption, and smog-induced visibility reduction can affect animal behaviours reliant on clear vision for hunting and navigation (Amann et al., 2017). Thus, extreme smog events not only pose significant health risks but also threaten environmental sustainability.

### c. Aerosol Index & its Role in Smog Forecasting

The aerosol index (AI) is a crucial metric for understanding and predicting air pollution, particularly in relation to smog formation. AI measures the concentration of airborne particulate matter, such as PM2.5 and PM10, which are essential components of smog. Smog, particularly in urban environments, results from a complex interaction between aerosols and other pollutants, such as nitrogen oxides (NOx) and volatile organic compounds (VOCs), under specific meteorological conditions. These pollutants react in the presence of sunlight to produce photochemical smog (Finlayson-Pitts & Pitts,

2000). Aerosols, being primary contributors to this reaction, play a pivotal role in the formation and intensity of smog.

The aerosol index is particularly useful in smog forecasting because it reflects the level of scattering and absorption of sunlight by aerosols, providing a direct measure of the particulate matter present in the atmosphere. High aerosol concentrations are often precursors to severe smog episodes, especially in densely populated and industrialised regions where emissions from traffic, factories, and agriculture are high (Bharali et al., 2023). As smog formation is highly dependent on both the concentration of pollutants and the atmospheric conditions, AI can serve as an early indicator of worsening air quality (Chen et al., 2022).

Using AI for smog forecasting is valuable because it allows for monitoring not only the overall levels of particulate matter but also the dynamic interactions between these particles and other pollutants. For instance, AI can provide insight into the transportation and accumulation of particulate matter, which are critical for understanding how and where smog will form (Lelieveld et al., 2015).

### d. ConvLSTM for Aerosol Index Forecasting & its Link to Smog

ConvLSTM models represent a powerful approach to forecasting aerosol indices due to their ability to simultaneously handle spatial and temporal dependencies in data. Traditional LSTMs are good at capturing temporal dependencies but fail to leverage the spatial context of data. ConvLSTM addresses this gap by incorporating convolutional layers, which enable the model to extract spatial features (such as local aerosol concentration patterns) while maintaining a memory of temporal sequences. This dual capability makes ConvLSTM an ideal candidate for forecasting the movement and intensity of aerosol concentrations across regions over time (Shi et al., 2015). A detailed mathematical overview of the ConvLSTM architecture is available in the supplement to this paper.

Aerosol forecasting is directly linked to smog research due to the role aerosols play in the formation of smog. Smog particles are central to the aerosol index, meaning that AI forecasting offers a pathway to predicting smog events (Finlayson-Pitts & Pitts, 2000). Given the transient and localised nature of both aerosols and smog, the spatial and temporal forecasting power of ConvLSTM models allows for better prediction of smog episodes.

Studies have demonstrated the utility of machine learning in air pollution prediction, with ConvLSTM models being particularly effective in addressing spatiotemporal challenges (Liao et al., 2020). By accurately predicting aerosol distribution, ConvLSTM models can provide early warnings of smog formation, giving policymakers the opportunity to implement interventions, such as traffic restrictions or industrial slowdowns, to mitigate smog's harmful effects on public health as well as on the environment (Peng et al., 2023).

## 2. Methods

An arbitrary study area was chosen which was confined to a bounding box covering northern regions of South Asia as these regions continuously experience some of the world's worst air quality (Majeed et al., 2024). The bounding box coordinates were:

```
bbox = [68.137207,24.886436,84.836426,34.379713] #WGS84 // lon,lat,lon,lat
```

The time period for the study was chosen from 01.01.2019 to 31.12.2023. The data was downloaded from the Sentinel-5P data store using the Deep Earth System Data Lab's xcube sentinel datastore (Brandt, 2023). The data was downloaded in the form of netCDF4 files which contain the AI data for the study area and time period. The data was then pre-processed and split into training and testing datasets. The training dataset contains the AI data from 01.01.2019 to 31.02.2022 and the testing dataset contains the AI data from 01.01.2023 to 31.12.2023.

The spatial resolution of the Sentinel-5P data is 3.629km x 3.269km per pixel, which was downsampled using bilinear interpolation.

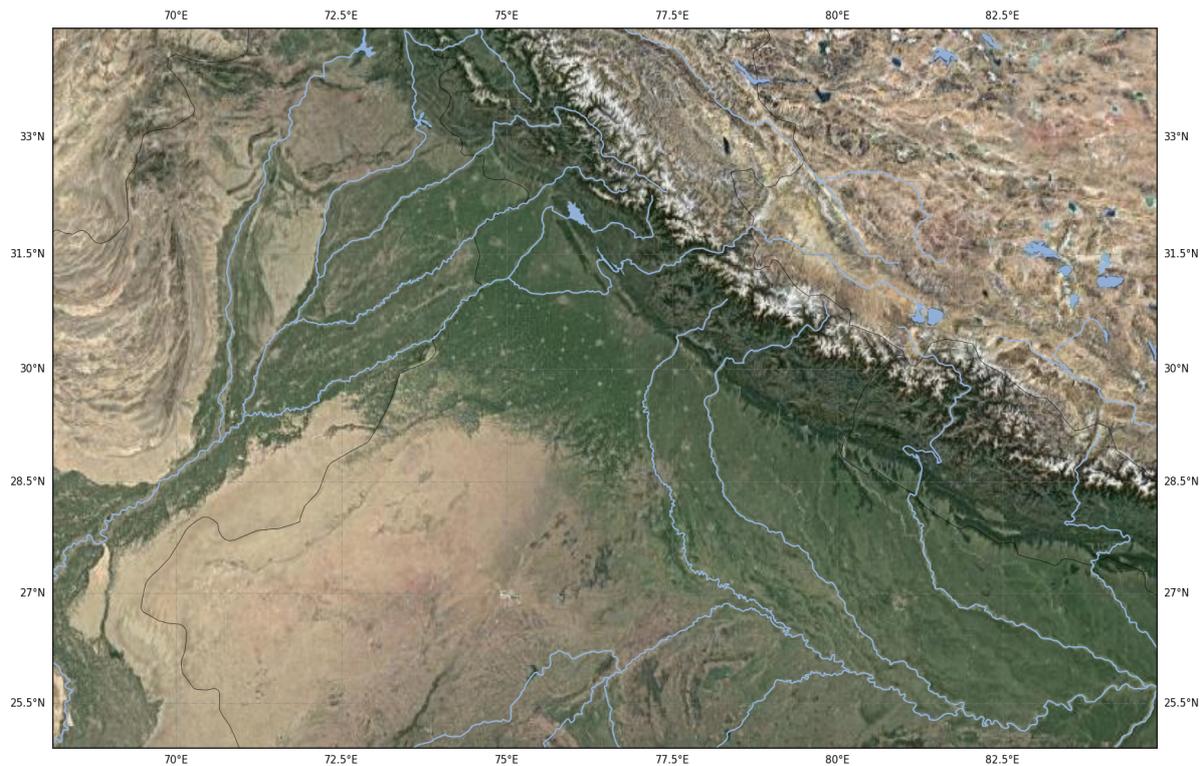

*Figure 1:* *Study area covering the Indo-Gangetic Plains.*

### a. Satellite Data Acquisition

I utilised an extensive dataset of atmospheric pollutants, specifically focusing on six key features: sulphur dioxide ($SO_2$), nitrogen dioxide ($NO_2$), methane ($CH_4$), ozone ($O_3$), carbon monoxide (CO), and formaldehyde (HCHO) to predict the target variable of Aerosol Index (AER_AI). These data were obtained from the Sentinel-5 Precursor (Sentinel-5P) satellite, part of the European Space Agency's Copernicus program (ESA, 2021). The satellite is equipped with the Tropospheric Monitoring Instrument (TROPOMI), which measures ultraviolet (UV) radiances at two specific wavelengths. The Sentinel-5P data was downloaded from the Earth System Data Lab's xcube Sentinel Data Store (Brandt et al., 2023).

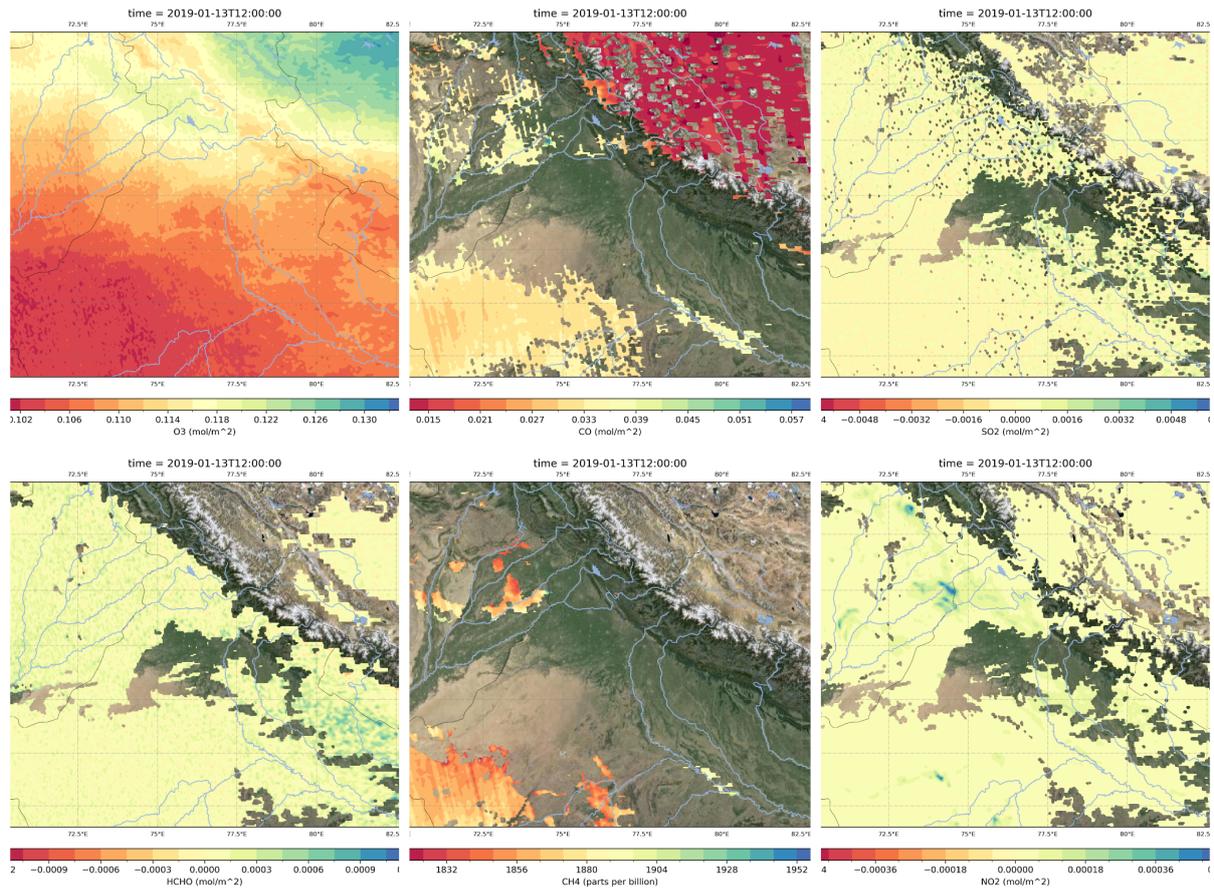

*Figure 2:* All six features (predictors) for the ConvLSTM model plotted over the study area for a single time (13-01-2019 at 12:00h). The predictors are (from top left): $O_3$, $CO$, $SO_2$, $HCHO$, $CH_4$, $NO_2$. All predictors apart from $CH_4$ are measured in $mol/m^2$. $CH_4$ is measured in parts per billion.

### b. Aerosol Index (AI)

The Aerosol Index is pre-calculated as a Sentinel-5p Level 2 product based on differences in backscattered UV radiation at two distinct wavelengths (ESA, 2021). The AI is sensitive to the presence of absorbing aerosols such as smoke, dust, and soot, as well as non-absorbing aerosols like sulphates. Positive AI values typically indicate the presence of absorbing aerosols, with higher values corresponding to more intense aerosol events. Negative and near-zero AI values may indicate non-absorbing aerosols or minimal aerosol presence, respectively.

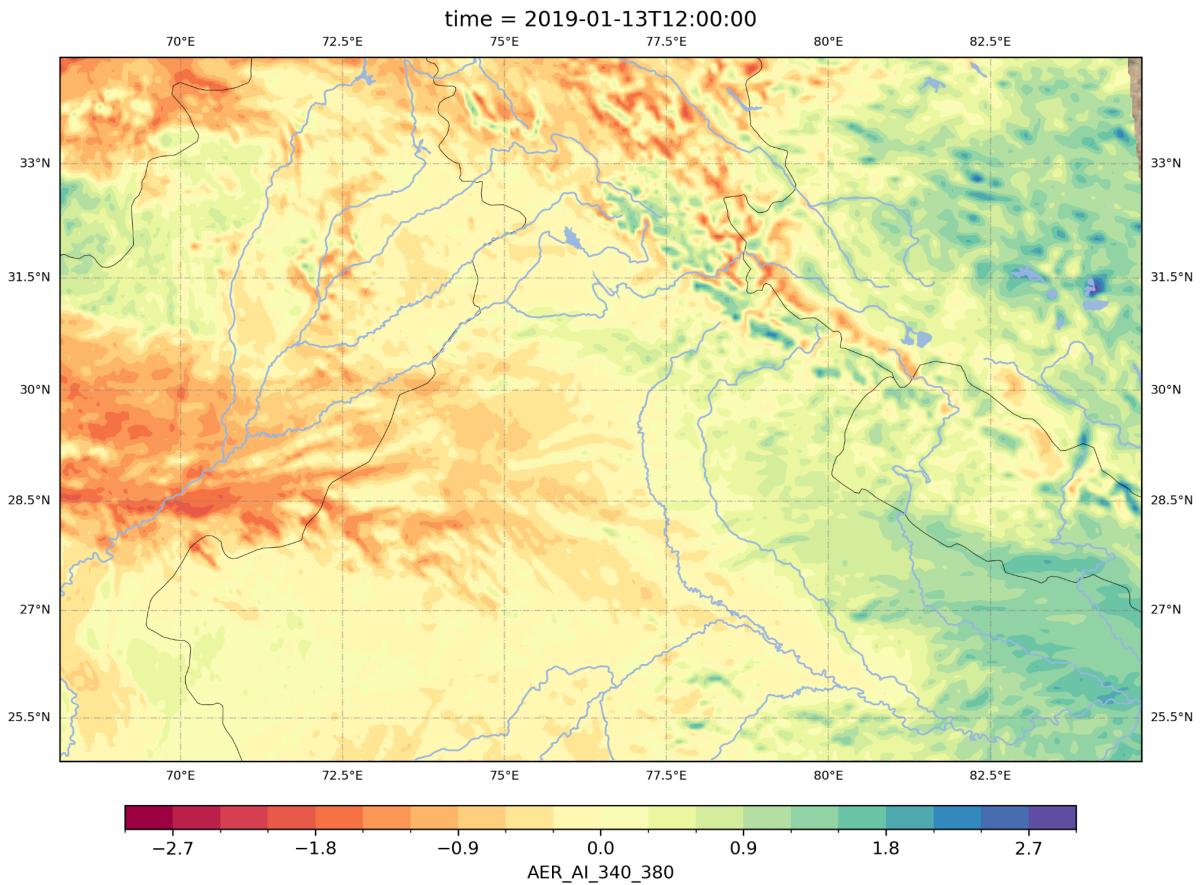

*Figure 3: Aerosol Index over the study area for 13-01-2019 at 12:00h.*

### c. Data Processing

Smog events, in particular, were identified by examining regions with persistent positive AI values, indicating the presence of absorbing aerosols such as soot and organic carbon. The spatial and temporal distribution of AI values was analysed to track the movement and intensity of smog over time.

Each feature (i.e. pollutant) is represented as a multidimensional array across (time, latitude, longitude, feature). Figure 4 shows the data processing steps that were applied. After imputation, the data was reshaped to include a timesteps dimension, resulting in an array of shape (samples, timesteps, latitude, longitude, features).

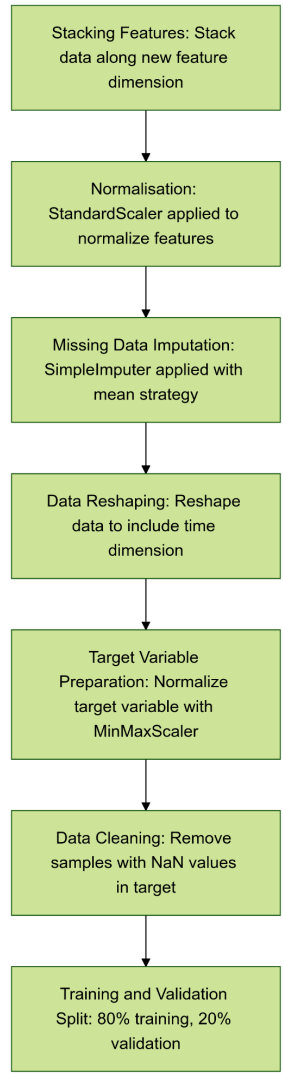

*Figure 4: An overview of the data processing steps in this study.*

### d. Model Architecture

Based on the original paper (Kumar et al., 2020), I developed a Convolutional Long Short-Term Memory (ConvLSTM) neural network (Khan, 2024). The architecture of the model summarised in Table 1 is as follows:

**Input Layer**: The input to the model was of shape (None, 1, latitude, longitude, features), where None represents the batch size.

**ConvLSTM2D Layers**: The first ConvLSTM2D layer consisted of 16 filters, each with a kernel size of 3x3, utilising tanh activation for output and sigmoid for recurrent activations. The layer was followed by batch normalisation to stabilise learning. A second ConvLSTM2D layer with 32 filters was added, maintaining the same kernel size and activations as the first layer.

**Conv3D Layer**: A 3D convolutional layer with 1 filter and a kernel size of 3x3x3 was employed as the final layer, with a sigmoid activation function to generate predictions in the range [0, 1].

**Model Compilation**: The model was compiled using the Adam optimizer with a learning rate of 1e-5, incorporating gradient clipping (clipnorm=1.0) to prevent exploding gradients. The loss function used was binary cross entropy, and accuracy was tracked as a performance metric.

| Layer (type) | Output Shape | Param # |
| --- | --- | --- |
| input_layer_1 (InputLayer) | (None, 1, 291, 512, 6) | 0 |
| batch_normalization_3 (BatchNormalization) | (None, 1, 291, 512, 6) | 24 |
| conv_lstm2d_2 (ConvLSTM2D) | (None, 1, 291, 512, 16) | 12,736 |
| batch_normalization_4 (BatchNormalization) | (None, 1, 291, 512, 16) | 64 |
| conv_lstm2d_3 (ConvLSTM2D) | (None, 1, 291, 512, 32) | 55,424 |
| batch_normalization_5 (BatchNormalization) | (None, 1, 291, 512, 32) | 128 |
| conv3d_1 (Conv3D) | (None, 1, 291, 512, 1) | 865 |

*Table 1*: An overview of the layers of the ConvLSTM model used in this study, where the input layer has dimensions of (None, 1, 291,512,6), and output layer of (None, 1, 291, 512, 1).

### e. Training Procedure

Training was conducted using a custom DataGenerator class to handle large datasets efficiently. The generator was designed to load data in batches, shuffling the data at the end of each epoch to ensure robust training.

**Hyperparameters**: Two models were trained for 50 and 100 epochs with a batch size of 1. Early stopping and learning rate reduction on plateau were implemented to prevent overfitting and adjust the learning rate dynamically.

**Validation and Monitoring**: Training progress was monitored using TensorBoard, capturing both training and validation losses and accuracies.

### f. Model Evaluation & Visualization

After training, the model's performance was assessed on the validation dataset. The predicted outputs were compared with the actual target values to compute validation loss and accuracy. The evaluation metrics used are:

**Binary Cross Entropy (Loss Function):** The loss function used in the model is binary cross entropy because the model output is binary (for example, predicting whether certain conditions in a given grid cell will exceed a particular threshold or not) (Goodfellow, 2016). The binary cross entropy loss is calculated as:

$$L(y, \hat{y}) = -\frac{1}{N} \sum_{i=1}^{N} \left[ y_i \cdot \log(\hat{y}_i) + (1 - y_i) \cdot \log(1 - \hat{y}_i) \right]$$

Where:
- $y_i$ is the true label (either 0 or 1).
- $\hat{y}_i$ is the predicted probability (between 0 and 1).

- $N$ is the number of samples.

**Mean Squared Error (MSE):** The mean squared error (MSE) measures the average of the squared differences between the predicted and true values (Goodfellow, 2016). It is used to track the model's performance for continuous regression outputs or to monitor general prediction error. The formula for MSE is:

$$MSE(y, \hat{y}) = \frac{1}{N} \sum_{i=1}^{N} (y_i - \hat{y}_i)^2$$

Where:
- $y_i$ is the true value.
- $\hat{y}_i$ is the predicted value.
- $N$ is the number of samples.

**Structural Similarity Index (SSIM):** is a metric used to assess the similarity between two images. Unlike traditional measures such as Mean Squared Error (MSE) or Pixel Difference, which simply compute the differences in pixel values, SSIM attempts to model the human visual system's perception of image quality (Wang et al., 2004). SSID is calculated for the validation images vs prediction images for the test dataset. The SSID is calculated with:

$$\text{SSIM}(x, y) = \left( \frac{2\mu_x \mu_y + C_1}{\mu_x^2 + \mu_y^2 + C_1} \right) \cdot \left( \frac{2\sigma_{xy} + C_2}{\sigma_x^2 + \sigma_y^2 + C_2} \right)$$

Where:
- $\mu_x$ and $\mu_y$ are the mean values (luminance) of the images $x$ and $y$, respectively.
- $\sigma_x^2$ and $\sigma_y^2$ are the variances (contrast) of the images.
- $\sigma_{xy}$ is the covariance (structural similarity) between the images.
- $C_1$ and $C_2$ are small constants added to avoid division by zero.

Based on these metrics, the following visualisations were created.

**i. Loss and Accuracy Curves**: Training and validation loss and accuracy over epochs plotted to assess the model's convergence.

**Ii. Prediction Plots**: Ground truth versus predicted values were visualised for a subset of validation samples to qualitatively assess the model's performance.

**Iii. Validation vs Predicted data plots:** Validation plotted again prediction for a single location and all timesteps.

**Iv. SSIM plots:** SSIM between all prediction and validation data (year 2023) for both epoch models plotted on the same line plot.

### g. Software & Tools

The entire workflow was implemented in Python, utilising libraries such as TensorFlow for neural network construction and training (Keras, 2015), Scikit-learn (Pedregosa et al., 2011) and xarray

(Hoyer & Hamman, 2017) for data preprocessing, and Matplotlib for visualisation. The model and training history were saved for future analysis and reproduction of results.

### h. Limitations

It is important to note that while the AI is effective in detecting and quantifying aerosols, it may have limitations in distinguishing between different aerosol types, especially in cases where aerosols have similar optical properties. Additionally, cloud cover and atmospheric conditions may affect AI accuracy, requiring careful interpretation of the results.

## 3. Results

The performance of the ConvLSTM model was evaluated by comparing its predicted aerosol images with the corresponding ground truth data (Figure 5). Across multiple validation examples, it was observed that while the model captured the broader spatial structures of the aerosol patterns, there was a noticeable smoothing effect in the predictions.

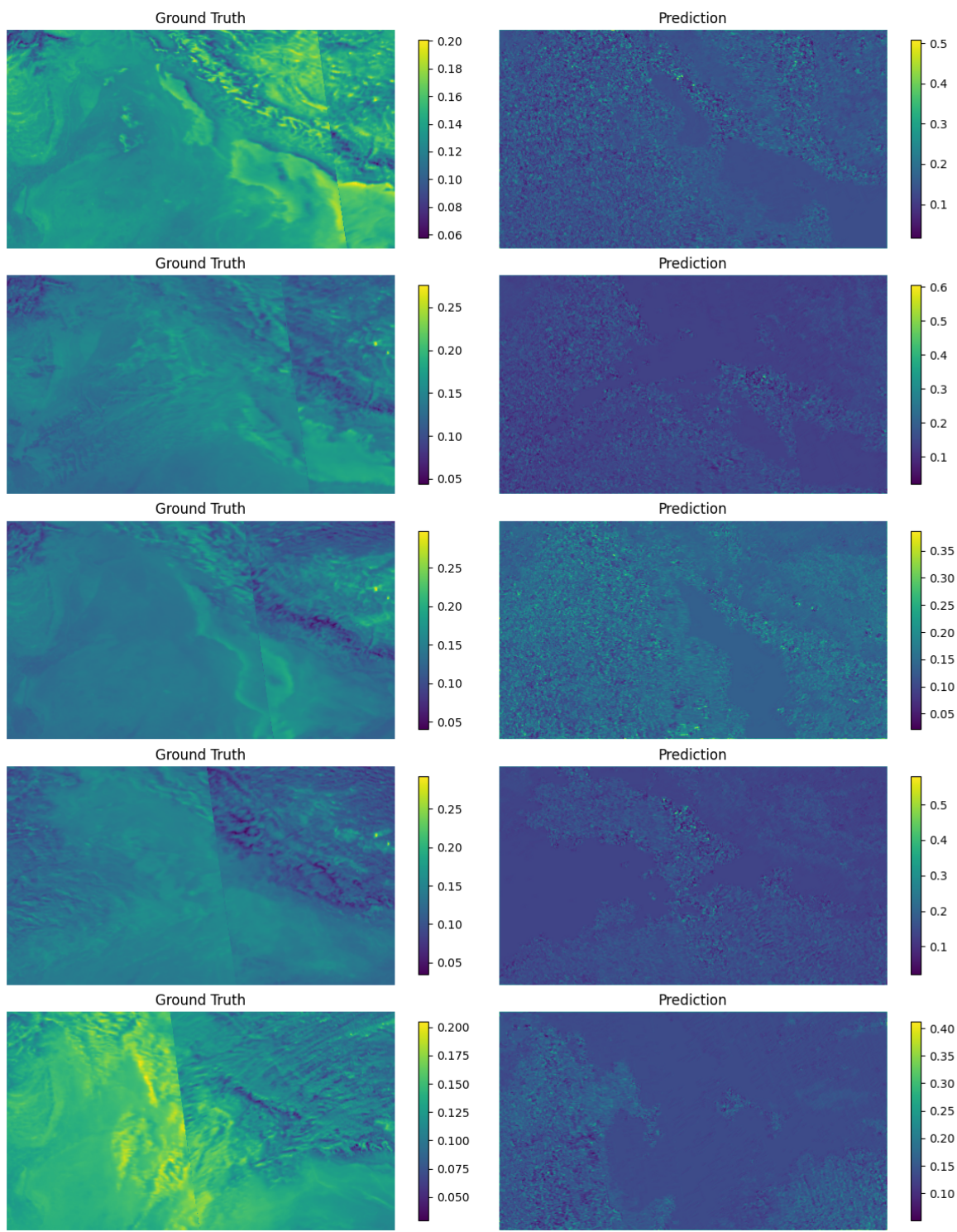

***Figure 5:*** *Visual comparison of ground truth (validation) data vs predictions of the 50 epoch model for 5 timesteps over the entire study area. These five figures show five timesteps starting from 01-01-2023 to 26-01.2023.*

### a. Training and Validation Loss & MSE over Epochs:

The training histories for both 50 and 100 epoch models reveal a steady decline in both the training and validation loss (left plot) and Mean Squared Error (MSE) (right plot) (Figures 6), indicating that the model is progressively learning and improving. Both the training and validation curves for loss and MSE follow similar trajectories, suggesting that the model generalises well to the validation data without significant overfitting. The final validation loss stabilises around 0.40, while the final validation MSE is approximately 0.03, indicating that the ConvLSTM model has captured the broader spatial-temporal trends but still struggles with finer detail prediction. The similarity in validation and training losses shows that the model is not overfitting but could be underfitting certain high-frequency components or intricate spatial patterns of aerosol distribution (Goodfellow et al., 2016).

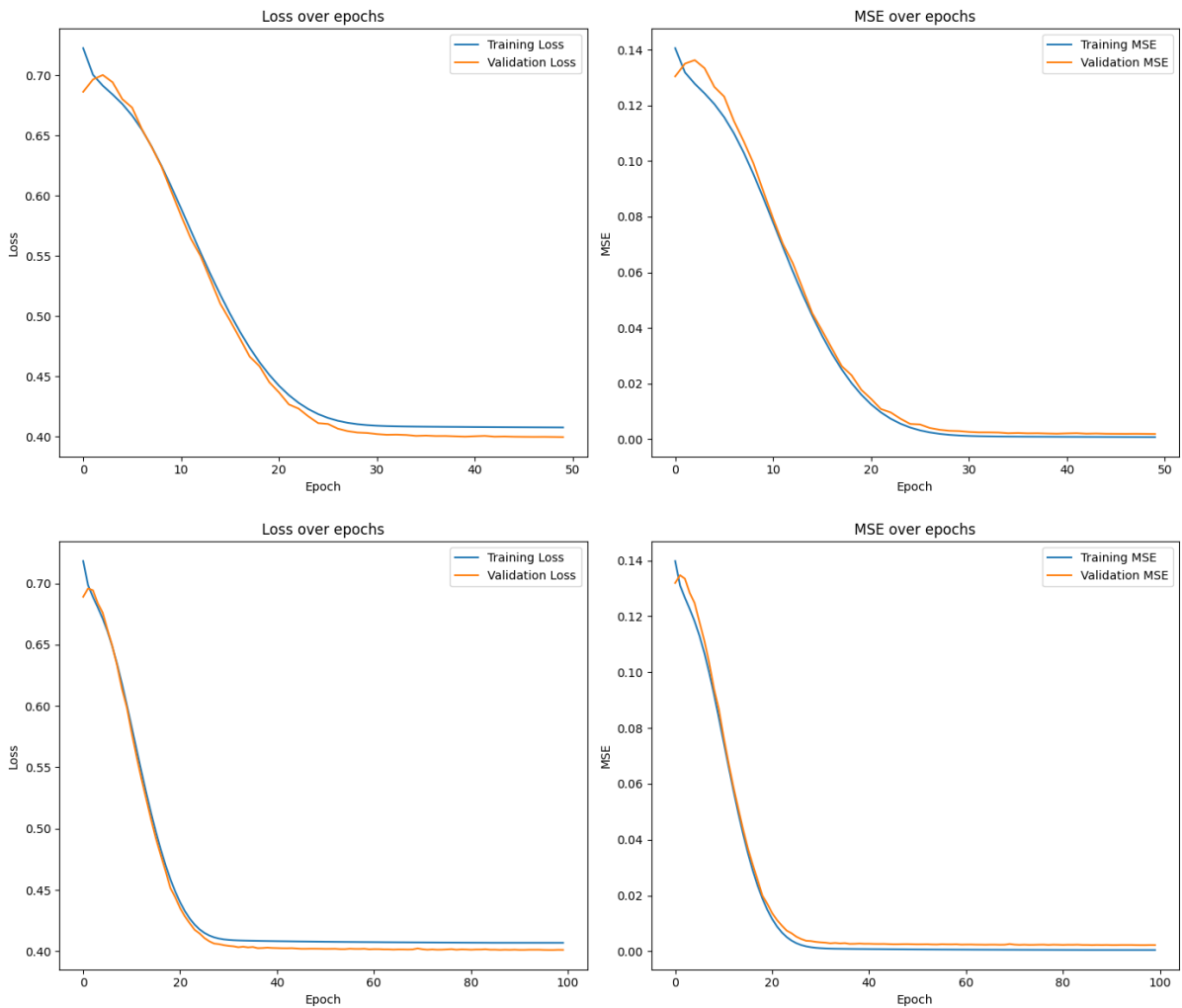

*Figure 6:* Training and validation loss (left) and Mean Squared Error (MSE) (right) over 50 (above) and 100 (below) epochs for the ConvLSTM model. Both metrics steadily decrease, with training and validation curves converging, indicating improved model performance and reduced overfitting as the number of epochs increases.

### b. Prediction vs Validation Data:

Figure 7 compares the predicted Aerosol Index (AI) values against the validation data over a period of 365 days for both models (50 and 100 epochs) for a single sample location, with the x-axis representing 5-day intervals. However, the full model is for 148992 locations. Varying levels of agreement between the predicted Aerosol Index (AI) and the ground truth values over the course of 365 days. The model predictions at both epochs generally capture the trend of the validation data, though the predictions exhibit notable fluctuations, particularly in the earlier timesteps. Predictions at epoch 50 show larger deviations from the validation data, suggesting that the model struggles more at this stage with capturing finer details. By epoch 100, the predictions show improved alignment, particularly in capturing the peaks and troughs of the Aerosol Index, indicating better learning of temporal dependencies. However, both epochs still show instances where the model either overestimates or underestimates certain features, suggesting that while the overall trend is being followed, refinement in capturing short-term variations could further enhance predictive accuracy, indicating the model's limited capacity in accurately predicting the true values of AI for smog events. The over-prediction of aerosol concentrations may stem from challenges in generalising to unseen data, common in models that do not fully capture complex aerosol behaviours (Xu et al., 2020).

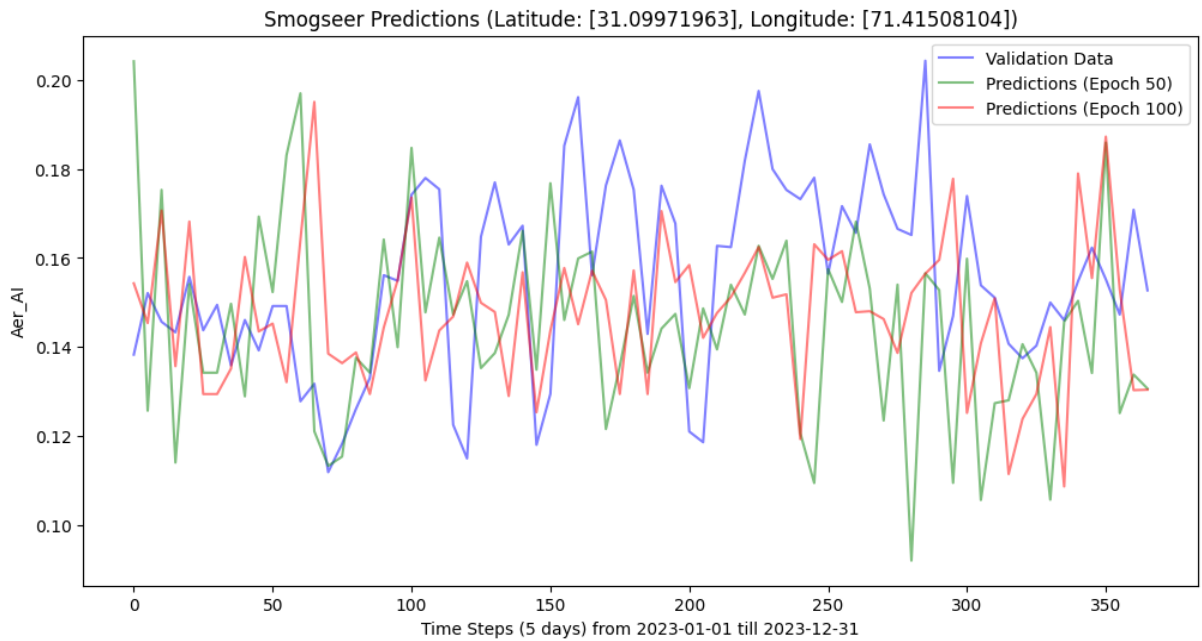

***Figure 7:*** *Comparison of validation data with ConvLSTM model predictions (epochs 50 and 100) for Aerosol Index at a single location, showing varying levels of prediction accuracy. The model follows the general trend but struggles with short-term fluctuations and high-frequency details.*

### c. Structural Similarity Index Measure (SSIM):

The SSIM values are a critical metric used to assess the similarity between the predicted AI maps and the actual ground truth data. In Figure 7, I compare the SSIM values at Epoch 50 (green line) and Epoch 100 (red line) across all time steps. SSIM values at Epoch 100 show higher values in the early time steps, signifying better performance at those points compared to Epoch 50. However, there is a noticeable drop in SSIM values for both epochs around the 200th timestep, which corresponds to a period where the model struggles to maintain structural consistency with the ground truth data. This drop could indicate that the model had difficulty capturing the spatial distribution patterns of aerosol concentrations during that time, possibly because of unmodeled dynamic changes in the atmospheric conditions. The overall improvement in SSIM from Epoch 50 to Epoch 100 suggests that additional training helps refine spatial accuracy, although diminishing returns are visible as training progresses (Zhang et al., 2019).

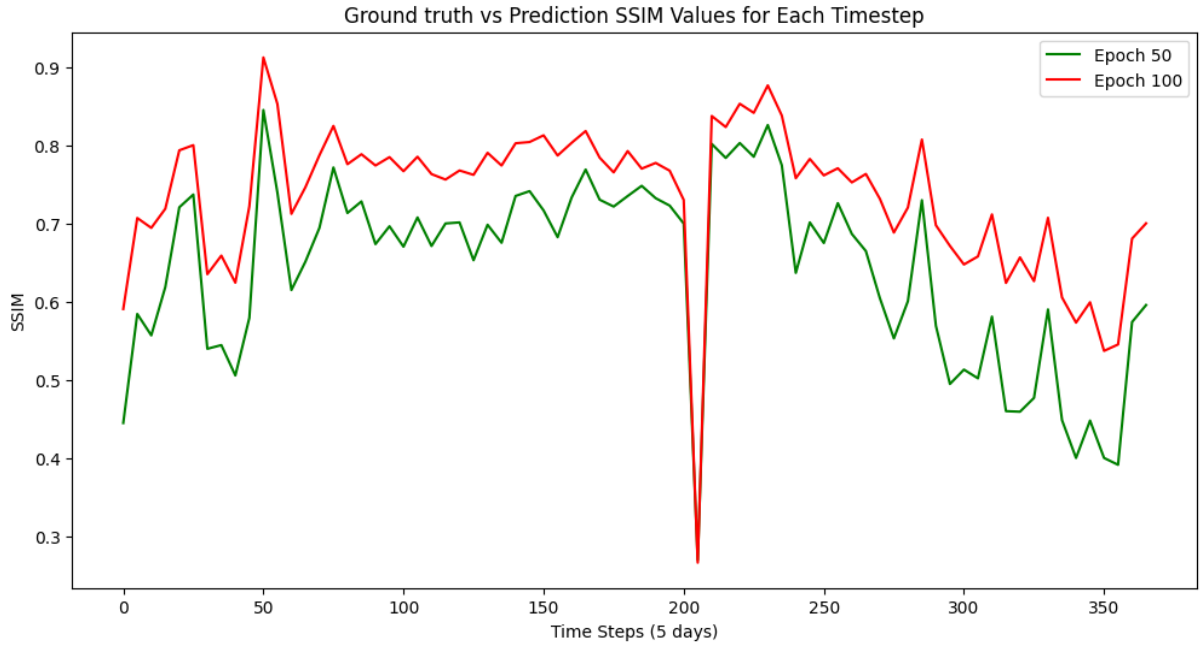

*Figure 7:* Structural Similarity Index (SSIM) values across time steps for model predictions at Epochs 50 and 100. The plot illustrates how SSIM scores, which measure the similarity between predicted and ground truth images, vary over time. Notably, the model's performance improves with more training (Epoch 100), as evidenced by generally higher SSIM values, though some fluctuations still exist around time steps with higher variability in aerosol events.

### d. Evaluation

Table 2 shows the loss and MSE values of the models, as well as the average SSIM. While the ConvLSTM model demonstrates adequate predictive capabilities, especially in maintaining the broader structure of aerosol concentrations, the results also highlight areas for improvement. The consistent offset in predicted AI values, as seen in Figure 7, points to potential biases in the model's understanding of aerosol distributions. Additionally, while SSIM improves with further training, the significant drops at later timesteps suggest that model refinement is necessary to capture sudden dynamic changes in aerosol patterns. The loss and MSE curves show stable and converging behaviour, indicating that the model is well-calibrated, though additional fine-tuning might be required to improve its predictive accuracy, particularly for regions with extreme smog events.

| Epochs | Loss | MSE | Avg. SSIM |
| --- | --- | --- | --- |
| 50 | 0.39958614110946655 | 0.0018263210076838732 | 0.6412104046280084 |
| 100 | 0.40102294087409973 | 0.0022550118155777454 | 0.7355129262049926 |

**Table 2**: Summary of evaluation statistics for the forecasting models.

## 4. Discussion

In Figure 7, predictions from the ConvLSTM model at epochs 50 and 100 generally follow the trend of the validation data across 365 days. While both epochs capture some key patterns, significant deviations are observed, particularly at epoch 50, where the model struggles to match the validation data, especially in the earlier time steps. By epoch 100, the model demonstrates a better ability to predict both peaks and troughs, indicating an improved grasp of the temporal dynamics. However, instances where the model overestimates or underestimates the AI persist, showing that while long-term trends are reasonably followed, short-term fluctuations still challenge the model's predictive capability.

In particular, the model struggled to accurately predict regions of high aerosol density. The ground truth images displayed well-defined areas of sharp aerosol gradients, especially in central and peripheral regions, whereas the model's predictions lacked these sharp transitions (Figure 5). This suggests that the ConvLSTM model had difficulty capturing high-frequency components and fine-scale features, leading to overly smoothed results. The smoothing effect was evident across several examples, where the general spatial patterns were maintained, but important finer details were lost, potentially indicating over-regularization or challenges in generalising small-scale aerosol structures.

Additionally, some prediction outputs showed signs of visual noise that were absent from the ground truth data (Figure 5). This noise might reflect the model's struggle to approximate complex or uncertain aerosol features in the validation data. Despite retaining the general distribution of aerosol patterns, the ConvLSTM model's limitations in capturing finer-scale features and high-intensity gradients highlight the need for further improvements. These enhancements could involve refining the model architecture or exploring alternative training techniques to improve its ability to predict more detailed aerosol structures.

The results of the ConvLSTM model in predicting aerosol patterns can be significantly improved by linking them to existing smog forecasting research. Smog, which is primarily composed of fine particulate matter (PM2.5) and other aerosols, shares many characteristics with the aerosol data used in this study. Traditional smog forecasting models, including chemical transport models (CTMs) and statistical approaches, have encountered difficulties in capturing fine-scale features and sharp aerosol gradients, particularly in regions with high aerosol density (Chen et al., 2022). These models often smooth over the spatial variability of pollutant concentrations, leading to inaccuracies in predicting peak pollution levels (Wang, 2020; Bocquet et al., 2015). Similarly, in the ConvLSTM model, I observed that while it could capture the broader spatial structures of aerosol patterns, the model produced a noticeable smoothing effect, especially in regions of high aerosol density. These findings align with existing challenges in smog research, where models tend to underestimate extreme pollution levels due to over-smoothing (Bocquet et al., 2015).

The ConvLSTM model's difficulty in capturing fine details, such as sharp aerosol gradients and localised high-intensity regions, mirrors the struggles faced in smog forecasting models. Research in smog prediction has highlighted similar limitations, particularly in forecasting fine-scale aerosol structures and maintaining sharp transitions in pollutant concentrations (Li et al., 2020; Tian et al., 2022; Stephens & Price, 1972). The observed smoothing in the ConvLSTM model's output could be attributed to its difficulty in representing high-frequency components of aerosol patterns, a common issue in models designed for atmospheric predictions. These limitations are crucial, especially when forecasting pollution levels in highly industrialised or urban areas where aerosol concentrations can exhibit significant spatial variability.

Recent advancements in smog forecasting have already explored the integration of deep learning techniques to address these challenges. Hybrid models that combine convolutional neural networks (CNNs) with recurrent neural networks (RNNs) have shown promise in improving the accuracy of smog forecasts, but they still encounter challenges in retaining high-resolution details in areas of peak pollution density (Li et al., 2020; Pan et al., 2023). These models have sought to capture the complex spatial and temporal dependencies in pollution data, but often face similar issues of over-smoothing and underestimation of localised pollution events. The ConvLSTM model's struggle with high-frequency features and fine aerosol structures aligns with these difficulties, suggesting that further improvements could be made by adopting strategies from recent smog prediction research.

For instance, multi-scale modelling approaches that integrate coarse and fine-resolution data have shown potential in improving the predictive performance of smog forecasting models. These methods allow models to capture both broad spatial trends and localised aerosol concentrations, which could help address the ConvLSTM model's limitations in predicting high-intensity aerosol regions (El-Harbawi, 2013). Additionally, incorporating external data sources, such as meteorological data and satellite observations, has been found to enhance the predictive capabilities of smog models, particularly in capturing spatial variability and fine-scale aerosol structures (Guttikunda & Gujjar, 2012). These strategies could be applied to improve the ConvLSTM model's ability to predict more detailed aerosol structures, which would be critical for accurate smog and air quality monitoring applications.

Another avenue for improvement could be refining the ConvLSTM model's architecture to better preserve high-frequency components and reduce over-smoothing. Techniques such as attention mechanisms or advanced loss functions that penalise the loss of fine-scale features might help improve the model's performance in capturing sharp aerosol gradients and localised high-density regions (Tao et al., 2020). By integrating these advancements from smog forecasting research, the ConvLSTM model could become more robust in predicting fine aerosol structures and high-intensity aerosol gradients, which are crucial for real-world applications in pollution forecasting and air quality monitoring.

In summary, the results of the ConvLSTM model, particularly its smoothing effects and difficulty in predicting high-density aerosol regions, are reflective of challenges encountered in smog forecasting research. By linking the model's limitations to these broader research efforts and adopting advancements from smog prediction models, such as multi-scale modelling and the integration of additional data sources, the predictive performance of the ConvLSTM model can be significantly improved. These improvements would enhance its utility in air quality monitoring and environmental research.

## 5. Conclusions

In this study, I evaluated the performance of a ConvLSTM model in predicting aerosol distribution by comparing its predictions with ground truth data. The results indicated that while the model effectively captured the broader spatial patterns of aerosol concentrations, it struggled to retain fine details, particularly in regions of high aerosol density. The smoothing effect observed in the model's outputs, along with the lack of sharp transitions in aerosol gradients, suggests limitations in the model's ability to capture high-frequency components and fine-scale features. These shortcomings were reflected across multiple validation examples, highlighting the need for improvement in preserving localised aerosol structures.

Linking my findings to research on smog forecasting models, which face similar challenges in retaining fine-scale pollutant features, further emphasised the relevance of addressing these limitations. Drawing from advances in smog prediction, such as multi-scale modelling, the integration of additional data sources like meteorological inputs, and refinements in model architecture, could provide potential pathways for enhancing the ConvLSTM model's performance. Incorporating these techniques may enable the model to better capture both broad spatial patterns and detailed aerosol structures, making it more robust for applications in air quality monitoring and smog prediction.

In conclusion, while the ConvLSTM model demonstrated promise in predicting aerosol patterns, future work should focus on improving its ability to capture finer-scale features and reducing the smoothing of high-intensity aerosol regions. By leveraging insights from smog forecasting research and exploring architectural enhancements, the model's utility in environmental monitoring and pollution prediction can be significantly enhanced. These improvements will be crucial for developing more accurate, high-resolution aerosol forecasting models that can contribute to better air quality management and public health protection.

# 7. Supplement

### a. ConvLSTM Neural Networks

A Convolutional Long Short-Term Memory (ConvLSTM) model is a specialised neural network architecture that integrates the strengths of Convolutional Neural Networks (CNNs) and Long Short-Term Memory networks (LSTMs) (Shi et al., 2015). This hybrid model is particularly well-suited for tasks involving spatiotemporal data, where both spatial and temporal dependencies are critical. ConvLSTM models are used extensively in fields such as video processing, weather forecasting, and environmental monitoring, where data exhibits strong correlations across both space and time (Hu et al., 2022).

The mathematical representation of a Convolutional Long Short-Term Memory (ConvLSTM) network combines the standard LSTM cell equations with convolution operations. It is particularly suited for spatiotemporal data, where the input has spatial dimensions (height and width) in addition to the temporal dimension. The primary difference between a ConvLSTM and a standard LSTM is that ConvLSTM replaces matrix multiplications with convolutions, making it more suitable for image sequences or other structured grid data.

Here's the mathematical representation of a ConvLSTM:

Let:

- $\mathbf{X}_t \in \mathbb{R}^{H \times W \times C}$ be the input at time step $t$, where $H$ is height, $W$ is width, and $C$ is the number of channels.
- $\mathbf{H}_t \in \mathbb{R}^{H \times W \times C}$ be the hidden state at time step $t$.
- $\mathbf{C}_t \in \mathbb{R}^{H \times W \times C}$ be the cell state at time step $t$.
- $*$ denotes the convolution operator, and $\sigma$ is the sigmoid activation function.
- $\odot$ is the Hadamard product (element-wise multiplication).

At each time step $t$, the ConvLSTM cell updates are computed as follows:

1. **Input Gate**:

$$\mathbf{i}_t = \sigma(\mathbf{W}_{xi} * \mathbf{X}_t + \mathbf{W}_{hi} * \mathbf{H}_{t-1} + \mathbf{b}_i)$$

This gate determines how much of the new input should be added to the cell state.

2. **Forget Gate**:

$$\mathbf{f}_t = \sigma(\mathbf{W}_{xf} * \mathbf{X}_t + \mathbf{W}_{hf} * \mathbf{H}_{t-1} + \mathbf{b}_f)$$

The forget gate decides how much of the previous cell state should be forgotten.

3. **Cell State Update**:

$$\tilde{\mathbf{C}}_t = \tanh(\mathbf{W}_{xc} * \mathbf{X}_t + \mathbf{W}_{hc} * \mathbf{H}_{t-1} + \mathbf{b}_c)$$

The candidate cell state $\tilde{\mathbf{C}}_t$ is generated using the current input and the previous hidden state.

$$\mathbf{C}_t = \mathbf{f}_t \odot \mathbf{C}_{t-1} + \mathbf{i}_t \odot \tilde{\mathbf{C}}_t$$

The new cell state $\mathbf{C}_t$ is a combination of the forget gate, the previous cell state, the input gate, and the candidate cell state.

4. **Output Gate**:

$$\mathbf{o}_t = \sigma(\mathbf{W}_{xo} * \mathbf{X}_t + \mathbf{W}_{ho} * \mathbf{H}_{t-1} + \mathbf{b}_o)$$

The output gate controls how much of the cell state should be passed to the hidden state.

5. **Hidden State Update**:

$$\mathbf{H}_t = \mathbf{o}_t \odot \tanh(\mathbf{C}_t)$$

The new hidden state $\mathbf{H}_t$ is computed based on the output gate and the current cell state.

**Summary of gates:**

- $\mathbf{i}_t$: Input gate controls the flow of new input.
- $\mathbf{f}_t$: Forget gate controls what part of the previous memory to retain.
- $\tilde{\mathbf{C}}_t$: Candidate cell state.
- $\mathbf{C}_t$: Current cell state after combining forget and input.
- $\mathbf{o}_t$: Output gate determines what information to pass to the next hidden state.

The key distinction between ConvLSTM and a fully connected LSTM lies in the convolutional operations (denoted by $*$) that are applied to the spatial data. This allows ConvLSTM to capture both temporal dependencies and spatial features, which is crucial for tasks such as video prediction or spatiotemporal sequence modelling